\pdfoutput=1
\documentclass[11pt]{article}
\usepackage[final]{acl}

\usepackage{times}
\usepackage{latexsym}
\usepackage{cleveref}
\usepackage{float}

\usepackage{adjustbox}
\usepackage{multirow}
\usepackage{graphicx}
\usepackage{makecell}
\usepackage{algorithm}
\usepackage{algpseudocode}
\usepackage[bottom]{footmisc}

\algnewcommand\algorithmicforeach{\textbf{for each}}
\algdef{S}[FOR]{ForEach}[1]{\algorithmicforeach\ #1\ \algorithmicdo}

\graphicspath{ {img/} }

\usepackage[T1]{fontenc}

\usepackage[utf8]{inputenc}

\usepackage{microtype}

\title{SKILL: Structured Knowledge Infusion for Large Language Models}

\author{
    Fedor Moiseev\thanks{\enskip Work done during internship at Google.}
    \hspace{0.3em}\textsuperscript{1,2} \quad 
    Zhe Dong\thanks{\enskip Correspondence Author.}
    \hspace{0.3em}\textsuperscript{2} \quad Enrique Alfonseca \textsuperscript{2} \quad Martin Jaggi 
    \textsuperscript{1} \\
    \textsuperscript{1}{EPFL, Switzerland} \quad \textsuperscript{2}{Google, Switzerland} \\
    \texttt{\small \{femoiseev, zhedong, ealfonseca\}@google.com,
    martin.jaggi@epfl.ch}
}

\begin{document}
\maketitle
\begin{abstract}
Large language models (LLMs) have demonstrated human-level performance on a vast spectrum of natural language tasks. However, it is largely unexplored whether they can better internalize knowledge from a structured data, such as a knowledge graph, or from text. In this work, we propose a method to infuse structured knowledge into LLMs, by directly training T5 models on factual triples of knowledge graphs (KGs). We show that models pre-trained on Wikidata KG with our method outperform the T5 baselines on FreebaseQA and WikiHop, as well as the Wikidata-answerable subset of TriviaQA and NaturalQuestions. The models pre-trained on factual triples compare competitively with the ones on natural language sentences that contain the same knowledge. Trained on a smaller size KG, WikiMovies, we saw $3\times$ improvement of exact match score on MetaQA task compared to T5 baseline. The proposed method has an advantage that no alignment between the knowledge graph and text corpus is required in curating training data. This makes our method particularly useful when working with industry-scale knowledge graphs.
\end{abstract}

\section{Introduction}
Large pre-trained language models, such as BERT \citep{devlin-etal-2019-bert}, GPT-3 \citep{gpt3}, T5 \citep{t5}, REALM \citep{realm} and ERNIE \citep{sun2021ernie} have become the state-of-the-art technology for many tasks. They are commonly pre-trained using unstructured text corpora, on tasks such as next word prediction, next sentence prediction (NSP) or masked language modelling (MLM). Especially for T5, self-supervised learning on unlabelled text corpus with MLM has been a common pre-training recipe \cite{roberts-etal-2020-much}. This is normally followed by a fine-tuning step on the task of interest \cite{ruder-etal-2019-transfer}, although large language models have also proved useful without this task-specific finetuning \cite{gpt3}.

Beyond the capacity of contextual understanding, human-level language understanding pivots on the knowledge about the world. The world knowledge is often expressed as factual triples \cite[c.f.][]{ji2020representation}, in the form of (\textit{subject entity}, \textit{relation}, \textit{object entity}). A knowledge graph (KG) defined by a set of factual triples consists of the subjects and objects as vertices/nodes, and the relations forming the edges connecting them. Most of the large scale KGs \cite[e.g. Wikidata,][]{wikidata} are stored in triple format.

LLMs demonstrate some capacity of learning world knowledge from the natural text corpus \cite{roberts-etal-2020-much}, but it is unclear to what degree they are also able to learn and memorize new knowledge directly from structured KG triples, or from text describing them explicitly.

In order to infuse knowledge into a LLM, one option is to generate a textual version of the knowledge base, and apply the standard training objectives, e.g. MLM. This is unfortunately highly nontrivial. One can either align sentences with KG triples, as done in ERNIE \cite{sun2021ernie}, or generate sentences from triples, as done in KELM \cite{agarwal-etal-2021-knowledge}. These approaches are unfortunately hard to port to knowledge graphs with different schemas. These processes are also lossy in that not every triple can be aligned or produce a valid sentence, and there is not a good understanding whether this can introduce unnecessary selection biases on top of biases existing in the original KG. 

In this work, we propose a method of \textbf{Knowledge Infusion for Large Language Models (SKILL)}, where LLMs directly learns from knowledge triples. Experiment results shows the checkpoints trained with proposed method on Wikidata KG outperform the T5 baselines on four standard closed-book question-answering (QA) tasks. With a smaller KG, WikiMovies, the proposed method gain $3\times$ exact match score performance improvement on MetaQA task. The models learning directly from knowledge triples performs competitively with the ones with the aligned natural sentences that contain the same amount of knowledge. Being  able  to  learn  directly from  knowledge triples enables easy addition of structured knowledge into language modeling pre-training.

\section{Related work}
Previous works that use knowledge graphs to enhance the quality of knowledge-intensive downstream tasks can be divided into two groups: using knowledge graphs at the inference time, and infusing knowledge into the model weights at the pre-training time. The proposed method falls in the latter group. 

\paragraph{Explicit usage of knowledge graphs.} 
A retrieval-augmented model is commonly used, in order to retrieve and apply the knowledge from external memories or sources. FILM \citep{verga-etal-2021-adaptable} and EaE \citep{fevry-etal-2020-entities} extend Transformer \citep{vaswani2017transformer} models with external entity (both FILM and EaE) and fact (FILM) memories. REALM \citep{realm} is pre-trained to perform reasoning over a large textual knowledge corpus on-the-fly during inference. UniK-QA \citep{oguz2020unik} combines the structured and unstructured information to improve the open-domain QA tasks with a retriever-reader framework. The main difference between the proposed method, SKILL, and retrieval-augmented models is that SKILL doesn't introduce retrieval system or external memories to the model, but it directly embeds knowledge into the model parameters, which introduces no extra cost at inference time.

\paragraph{Knowledge infusion.} 
A common way of parameterized knowledge infusion is to map or convert structured knowledges into natural language text. ERNIE 3.0 \citep{sun2021ernie} trains a knowledge-enhanced model on a corpus combining triples and their aligned sentences, by randomly masking relation in a triple or words in a sentence. On the contrary, SKILL trains only on triples.

KnowBert \citep{peters-etal-2019-knowledge} incorporates knowledge from Wikipedia and WordNet \citep{miller1995wordnet} into a BERT model through entity embeddings with knowledge-attention and re-contextualization mechanism. BERT-MK \citep{he-etal-2020-bert} is a BERT-based model that integrates graph contextual knowledge of a medical KG, which demonstrates the utility of graph-level knowledge. These approaches requires entity linking and sentences contextualizing the knowledge graph information.

KG-FiD \citep{yu2021kg} extends the Fusion-in-Decoder model \citep{izacard-grave-2021-leveraging}, with a module that filters and re-ranks passages based on structural connections in knowledge graph between entities described in those passages. In contrast to the SKILL method that we propose, it requires the existence of natural text passages describing each knowledge graph entity, so Wikipedia corpus was used since it naturally provides articles that describe entities.

\citet{heinzerling-inui-2021-language} explored the ability of language models to memorize and understand information from knowledge graphs, but used natural language representation of triples based on predefined templates instead of structured representation. Usage of predefined templates significantly limits scalability and therefore only relatively small knowledge graphs were used, such as Google-RE\footnote{https://ai.googleblog.com/2013/04/50000-lessons-on-how-to-read-relation.html}.

In contrast to the new method presented in this paper, all of these approaches require an explicit mapping between the knowledge graph entities or facts and corresponding natural language sentences, which can limit applications to industry-scale knowledge graphs that don't have such a mapping.

\paragraph{Different goals of using knowledge graphs.}
Besides that, some papers embed knowledge into model weights but pursue different goals rather than improving performance on downstream tasks. COMET \citep{bosselut-etal-2019-comet} is most similar to our work and trains a commonsense-aware Transformer Language Model by learning to generate loosely structured commonsense descriptions in the natural language given the structured knowledge. Similar to us, it also uses KG triples in surface form as a source for training data, but in contrast to our research, the final goal of COMET is to generate new knowledge instead of utilizing existing ones. Another important difference is the scale: COMET uses Atomic \citep{sap2019atomic} and ConceptNet \citep{speer2017conceptnet} Knowledge Graphs that are much smaller than Wikidata \citep{wikidata}.

KELM \citep{agarwal-etal-2021-knowledge} fine-tunes a T5 model to convert KGs to synthetic natural language sentences to augment existing pre-training corpora. We build our research on top of it and use the KELM dataset to compare structured and natural language representations of knowledge.

\begin{table*}
\centering
\begin{adjustbox}{center}
{\footnotesize
\begin{tabular}{ccccc}
\hline
\textbf{Wikidata triple}                                  & \textbf{KELM sentence}                                         & \textbf{Wikidata input}                    & \textbf{KELM input}                                            & \textbf{Target} \\ \hline
\makecell{("Pulp Fiction",\\ "award received",\\ "Palme d'Or")} & \makecell{Quentin Tarantino \\won the Palme d'Or in 1994\\ for Pulp Fiction.} & \makecell{Pulp Fiction,\\ award received,\\ {[}MASK{]}} & \makecell{Quentin Tarantino \\won the {[}MASK{]} in 1994\\ for Pulp Fiction.} & Palme d'Or      \\ \hline
\end{tabular}}
\end{adjustbox}
\caption{
\label{table:inputs}
Example inputs for SKILL pre-training with Wikidata and KELM corpora.
}
\end{table*}

\section{Method}
\label{sec:method}
There are two components of knowledge infusion for LLMs (SKILL): the corpus and the training method. We introduce the method based on Wikidata KG, but it can be applied to any other KGs.

\paragraph{Training corpus.} 
We use two corpora with different knowledge representations: Wikidata KG \cite{wikidata} in triple format, and KELM corpus\footnote{Data is available at https://github.com/google-research-datasets/KELM-corpus} \citep{agarwal-etal-2021-knowledge} as synthetic natural language sentences converted from Wikidata KG. The KELM corpus contains $15,628,486$ synthetic sentences. To ensure two corpora share the same knowledge, we take the snapshot of the Wikidata KG used to created the KELM corpus, which contains $35,697,715$ triples.

To prevent the degradation of model performance on natural language understanding, we mix the Wikidata corpus or KELM corpus with natural text from C4 \cite{t5}, $50:50$, for the knowledge infusion training data.

\paragraph{Training method.} 
T5 \cite{t5} was trained through masked-language modelling with random span corruption on the C4 corpus. \citet{roberts-etal-2020-much} found that masking salient terms \cite{realm} in pre-training T5 models, instead of masking random token spans, could significantly improve the performance on downstream tasks, e.g. closed-book QA. 

We apply salient span masking for unsupervised learning in our knowledge-infusing training. To mask the same amount of information is for both corpora, the following method is applied. For a knowledge triple, we mask either the subject or object entity. For a KELM sentence, we identify the aligned triple, with details in \Cref{sec:kelm-matching}, and mask the full spans corresponding to the subject or object in the triple. The \textit{relation} tokens are never masked, as there is no robust way to map the abstract relation in knowledge triples to natural language tokens in KELM sentences. Examples of the inputs for both corpora are in \Cref{table:inputs}.

\section{Experiments}
We assess SKILL by training and evaluating the knowledge infused models on closed-book QA tasks, where questions are provided without supporting context and external knowledge. 

\subsection{Experiment Setup}
\paragraph{SKILL pre-training.}
We apply SKILL on three T5.1.1 pre-trained checkpoints\footnote{https://goo.gle/t5-checkpoints}, base, large, and XXL, with sizes of $\sim 250$M, $\sim 800$M and $\sim11$B parameters, respectively. For T5.1.1-base and -large, SKILL training is performed for $500$K steps with batch size $1024$, which translates to $\sim 7.17$ epochs on Wikidata KG and $\sim 16.38$ epochs in KELM sentences. For T5.1.1-XXL, the model is trained for $100$K steps to finish training in a feasible time. 

As baseline we use pre-trained T5 checkpoints of the same size. To make sure that improvements come from knowledge infusion instead of from longer C4 pre-training, we use a second baseline by further training the T5 checkpoints on C4 for half of the aforementioned steps, to match the amount of C4 pre-training used in SKILL. 

All the model variations are optimized by AdaFactor \cite{adafactor} with $10^{-3}$ learning rate and  $0.1$ dropout rate, the same settings that were used for T5.

\paragraph{Fine-tuning on closed-book QA tasks.}
We evaluate the checkpoints by fine-tuning on the following QA benchmarks: FreebaseQA \citep{jiang-etal-2019-freebaseqa}, WikiHop \citep{welbl-etal-2018-constructing}, TriviaQA \citep{joshi-etal-2017-triviaqa} and NaturalQuestions \citep{kwiatkowski-etal-2019-natural}, with the aforementioned hyper-parameters for optimization and $128$ batch size. For the benchmarks without a \textit{test} split, we use the \textit{dev} split for test, and the last $10\%$ of \textit{train} as \textit{dev} split.

The Exact Match (EM) scores on the test sets are calculated after being fine-tuned for $50$K steps for T5.1.1-base and -large models, and $10$K steps for -XXL models. All models converged with no noticeable over-fitting according to the EM scores on validation sets.

\paragraph{Wikidata-answerable QA.}
\label{sec:wikidata-answerable-qa}

\begin{table*}[!ht]
\centering
\adjustbox{max width=\textwidth}{
\begin{tabular}{lcccccccccccc}
\hline
\textbf{Model} & \multicolumn{2}{c}{\textbf{FreebaseQA}}                              & \multicolumn{2}{c}{\textbf{WikiHop}}                                 & \multicolumn{2}{c}{\textbf{TQA-matched}}                             & \multicolumn{2}{c}{\textbf{TQA}}                                     & \multicolumn{2}{c}{\textbf{NQ-matched}}                              & \multicolumn{2}{c}{\textbf{NQ}}                                      \\
               & \multicolumn{1}{c}{\textbf{dev}} & \multicolumn{1}{c}{\textbf{test}} & \multicolumn{1}{c}{\textbf{dev}} & \multicolumn{1}{c}{\textbf{test}} & \multicolumn{1}{c}{\textbf{dev}} & \multicolumn{1}{c}{\textbf{test}} & \multicolumn{1}{c}{\textbf{dev}} & \multicolumn{1}{c}{\textbf{test}} & \multicolumn{1}{c}{\textbf{dev}} & \multicolumn{1}{c}{\textbf{test}} & \multicolumn{1}{c}{\textbf{dev}} & \multicolumn{1}{c}{\textbf{test}} \\ \hline
base           & $25.24$            & $27.55$            & $19.09$           & $18.38$          & $31.24$             & $33.55$            & $22.64$          & $22.93$          & $36.64$            & $32.68$            & $25.04$          & $25.48$          \\
base + C4      & $26.19$            & $28.33$            & $19.57$           & $19.36$          & $32.9$              & $34.4$             & $24.54$          & $25.39$          & $36.98$            & $32.03$            & $\mathbf{25.88}$ & $25.84$          \\
base + WikiKG      & $\mathbf{26.92}$   & $\mathbf{28.38}$   & $20.28$           & $\mathbf{20.22}$ & $\mathbf{34.21}$    & $35.08$            & $24.73$          & $\mathbf{25.77}$ & $\mathbf{37.41}$   & $\mathbf{33.33}$   & $25.51$          & $25.76$          \\
base + KELM    & $26.64$            & $28.15$            & $\mathbf{20.62}$  & $19.81$          & $33.64$             & $\mathbf{35.54}$   & $\mathbf{25.22}$ & $25.75$          & $36.98$            & $32.9$             & $25.31$          & $\mathbf{26.2}$  \\ \hline
large          & $30.22$            & $32.88$            & $20.92$           & $21.12$          & $36.7$              & $38.09$            & $29.24$          & $30.03$          & $39.22$            & $35.06$            & $27.12$          & $27.15$          \\
large + C4     & $32.55$            & $34.01$            & $22.5$            & $21.51$          & $38.78$             & $40.6$             & $30.32$          & $30.83$          & $39.74$            & $35.5$             & $27.46$          & $28.17$          \\
large + WikiKG     & $\mathbf{33.22}$   & $\mathbf{35.29}$   & $\mathbf{23.5}$   & $\mathbf{23.4}$  & $39.19$             & $\mathbf{41.02}$   & $29.74$          & $30.47$          & $\mathbf{41.12}$   & $\mathbf{35.93}$   & $27.38$          & $27.89$          \\
large + KELM   & $32.65$            & $34.16$            & $23.34$           & $22.91$          & $\mathbf{39.45}$    & $40.76$            & $\mathbf{30.51}$ & $\mathbf{30.65}$ & $40.95$            & $35.5$             & $\mathbf{27.67}$ & $\mathbf{28.56}$ \\ \hline
XXL            & $43.67$            & $45.02$            & $24.76$           & $24.8$           & $51.73$             & $53.1$             & $42.44$          & $42.21$          & $46.47$            & $43.72$            & $31$             & $32.27$          \\
XXL + C4       & $42.01$            & $44.14$            & $23.34$           & $22.23$          & $50.59$             & $52.19$            & $40.66$          & $40.99$          & $45.43$            & $40.26$            & $30.35$          & $31.08$          \\
XXL + WikiKG       & $45.22$            & $\mathbf{47.25}$   & $\mathbf{27.57}$  & $\mathbf{27.65}$ & $\mathbf{54.17}$    & $54.18$            & $42.55$          & $\mathbf{43.54}$ & $\mathbf{49.14}$   & $\mathbf{44.37}$   & $31.11$          & $\mathbf{32.74}$ \\
XXL + KELM     & $\mathbf{45.42}$   & $45.9$             & $26.11$           & $26.26$          & $53.65$             & $\mathbf{54.21}$   & $\mathbf{42.68}$ & $42.95$          & $48.53$            & $44.16$            & $\mathbf{31.79}$ & $32.6$                                     \\ \hline
\end{tabular}
}
\caption{\label{table:wikidata}
Exact match scores achieved by fine-tuning the checkpoints on closed-book QA tasks. \texttt{base}, \texttt{large}, \texttt{XXL} represent the corresponding T5.1.1-* checkpoints. \texttt{*-C4} are the checkpoints additionally trained on C4 corpus as discussed in \Cref{sec:method}. \texttt{*-WikiKG} and \texttt{*-KELM} are the checkpoints trained on Wikidata KG triple corpus and KELM sentence corpus. The best performed checkpoints are in bold. Details about datasets are in \Cref{sec:dataset-sizes}.}
\end{table*}

We found that the majority of the questions in FreebaseQA and WikiHop can be answered directly from triples in Wikidata. This is because FreebaseQA was created by matching question-answer pairs with triples in Freebase \citep{freebase}, most of which was imported into Wikidata \citep{wikidata}. For WikiHop, the questions were generated from Wikidata triples.  

However, TriviaQA and NaturalQuestions were created independently of Wikidata, and not every question can be answered using this knowledge base. We found frequent freshness issues, e.g. the golden answer for question "Who is the largest supermarket chain in the UK?" is "Aldi", while today it would be "Tesco". Some other questions can not be answered by WikiData, e.g. "Who, during a radio microphone test in 1984 said, 'I just signed legislation which outlaws Russia forever. The bombing begins in five minutes?'", with the golden answer "Ronald Reagan".  

To mitigate this, we created subsets of TriviaQA (TQA) and NaturalQuestions (NQ) that were somewhat more likely to have answers in Wikidata. We selected all the items for which there exist a triple in Wikidata that has the answer either as subject or object, and the other entity in the triple is mentioned in the question. We match the entities by entity name, case-insensitive. We name the Wikidata-aligned version of TQA and NQ as TQA-matched and NQ-matched, respectively. The dataset sizes of all QA tasks are summarized in \Cref{sec:dataset-sizes}.

\subsection{Results}
\label{sec:results}
The results for closed-book QA tasks are summarized in \Cref{table:wikidata}. SKILL pre-trained models show improvements on FreebaseQA, WikiHop, and Wikidata-answerable versions of TriviaQA and NaturalQuestions, but no significant improvement on original TriviaQA and NaturalQuestions. As discussed in previous section, we believe this is due to the misalignment between the datasets and Wikidata.

Models pre-trained on Wikidata KG gives competitive results with ones on KELM sentences. It shows that the triple representation is as good as natural language representation, while being much easier to scale up for larger KG.

For T5.1.1-base and -large, additional pre-training on C4 boosts performance in comparison to the original baseline. For T5.1.1-XXL, this additional pre-training leads to a performance regress. In \citep{t5}, it is mentioned that training on C4 for multiple times may reduce the performance of a T5 model. 

\begin{figure}
\centering
\includegraphics[width=0.85\columnwidth]{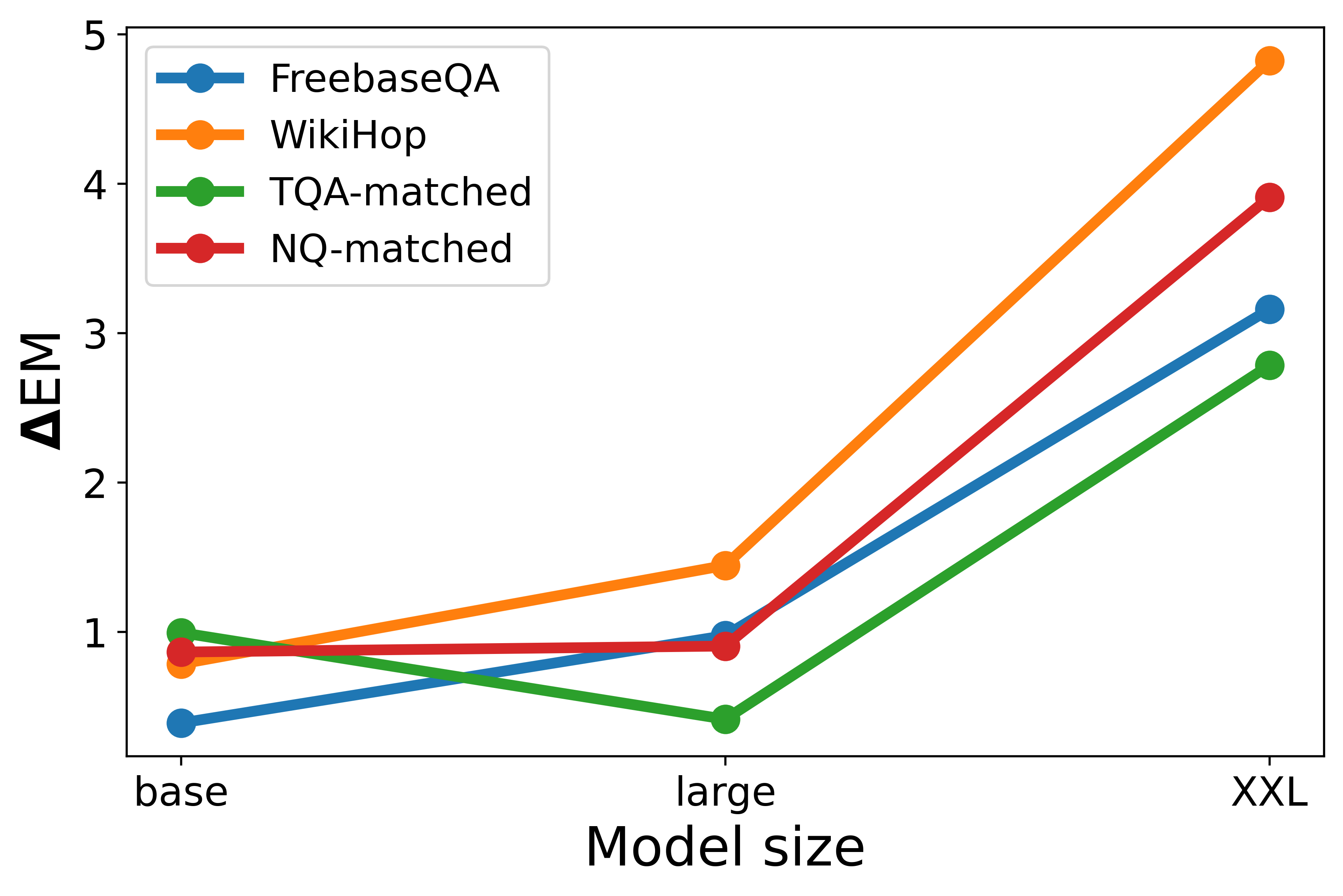}
\caption{\label{figure:delta-size}
Performance improvements on closed-book QA tasks for different model sizes. The improvements are measured by the difference of exact match score ($\Delta$EM) between knowledge-infused model trained with Wikidata triples and the baseline trained with C4 corpus. 
}
\end{figure}

\paragraph{Impact of model size.}
As shown in \Cref{figure:delta-size}, SKILL pre-training introduces bigger improvements when applied on larger models. With more than $35$M triples in Wikidata KG, it is harder for smaller size models, e.g. T5.1.1.-base with $300$M parameters, to memorize them efficiently. We view this as an encouraging result, suggesting that as model size grows, gains from SKILL pre-training may increase further.

\paragraph{Performance on a smaller KG.}
The WikiMovies KG \cite{miller-etal-2016-key} contains $134,741$ triples. T5.1.1-large should have enough parameters to memorize the KG. We train a T5.1.1-large model on the KG for $100$K steps, $\sim 380$ epochs, with the same hyperparameters as for Wikidata KG. We evaluate the checkpoints with MetaQA \citep{zhang2017variational} benchmark that was constructed over WikiMovies KG. The benchmark contains 3 different sub-tasks: 1-hop QA (e.g. "What films does Paresh Rawal appear in?"), 2-hop QA (e.g. "Who are the directors of the films written by Laura Kerr?"), 3-hop QA (e.g. "Who directed the movies written by the writer of Millennium Actress?").

The results in \Cref{table:metaqa} demonstrate the effectiveness of SKILL pre-training, when it's possible to memorize the whole knowledge graph. 

\begin{table}
\centering
\begin{adjustbox}{center}
{\footnotesize
\begin{tabular}{llccc}
\hline
 \textbf{Dataset}           &   \textbf{Split}   & \textbf{Baseline} & \textbf{+ C4} & \textbf{+ KG} \\
\hline
\multirow{2}{*}{1-hop} & dev  & $24.3$           & $23.12$         & $\mathbf{71.52}$         \\
                       & test & $24.5$           & $23.53$         & $\mathbf{71.47}$         \\
\multirow{2}{*}{2-hop} & dev  & $32.05$         & $32.23$         & $\mathbf{33.49}$         \\
                       & test & $32.65$          & $32.78$         & $\mathbf{33.57}$         \\
\multirow{2}{*}{3-hop} & dev  & $42.08$          & $39.22$         & $\mathbf{43.79}$         \\
                       & test & $42.31$          & $39.66$         & $\mathbf{43.41}$   \\
\hline
\end{tabular}}
\end{adjustbox}
\caption{\label{table:metaqa}
Exact match scores achieved by fine-tuning different T5.1.1-large checkpoints on MetaQA task.
}
\end{table}

As 1-hop questions are supported by single triples in the WikiMovies KG, a $3\times$ improvement on EM score is observed for the sub-task. In order to answer 2/3-hop questions it is not enough to memorize the triples, the model needs to be able to reason with them. This requires a better understanding of the graph structure. Training with single triples may not be enough, and the observed improvement is notably smaller. The performance could be further improved by representing more explicitly the graph structure in the training data, which we leave for future work.

\section{Conclusion}
We proposed a method to directly infuse knowledge from knowledge graphs into T5 models through pre-training. Empirical results show that T5 can learn directly from structured data and apply the learned knowledge to improve closed-book QA results. We also demonstrated that the models pre-trained on factual triples perform competitively with the ones on natural language sentences that contain the same knowledge. By enabling knowledge infusion directly from triples, this method can be very easily applied to industry-scale KGs.

\section{Ethical and Broader Impact}
In this work, we are introducing a new method to pre-train a well known natural language understanding model, T5, on the full corpora of public knowledge graphs. To the best of our knowledge, the method will not introduce extra bias to either the model or the dataset beyond the one potentially inherited from Wikidata \cite{wikidata} and WikiMovies \cite{miller-etal-2016-key} knowledge graphs. On the other hand, through knowledge fusion pre-training introduced in this work, a language model will be able to learn factual information to improve the quality of parameterized knowledge embedded in the model, which is demonstrated by improvements on various closed-book question-answering tasks. The proposed method and recipe will provide positive impact to the natural language processing community and help to improve the factualness in pre-trained large language model checkpoints.

\paragraph{Limitations.} A factual triple is the basic ingredient of a knowledge graph. However, as a semantic network, the graph structure of a knowledge graph describes how the factual triples are connected. This information is not easy to directly represent by random set of triples. We leave the exploration of how to infuse the information implied by the graph structure for future work. We expect that this will further improve the results, especially for multi-hop question-answering tasks.

\bibliography{anthology,custom}

\appendix

\section{Matching of entities in KELM sentences}
\label{sec:kelm-matching}
To find Wikidata KG entities in corresponding KELM sentences, we use \Cref{alg:kelm-matching}. Additional cycle on line \ref{line:fuzzy-matching} is needed because some entities have an information in brackets that should not be in a sentence, for example \texttt{John Doe (born 1990)}. This algorithm  matched at least one entity to $15,383,248$ out of $15,628,486$ KELM sentences.

\begin{algorithm}
\begin{algorithmic}[1]

\State $KELM_{matched} \gets \emptyset$
\ForEach {$k \in KELM$ sentences}
\ForEach {$t \in triples(k)$}
\ForEach {$e \in entities(t)$}
    \State $e_{p} \gets$ \Call{Preprocess}{$e$}
    \State $k_{p} \gets$ \Call{Preprocess}{$k$}
    \State $spans \gets$ \Call{MatchEntity}{$e_{p}$, $k_{p}$}
    \State $KELM_{matched}.insert([k, spans])$
\EndFor
\EndFor
\EndFor
\State
\Function{MatchEntity}{$e$: entity, $k$: KELM sentence}
\State $spans \gets \emptyset$
\ForEach{$s \subset k: date(e) = date(s)$}
    \State $spans.insert(s)$
\EndFor
\ForEach{$\exists s \subset k: e = s$}
    \State $spans.insert(s)$
\EndFor
\If {$spans = \emptyset$}
\ForEach{$\exists s \subset k: e = s+$" (*)"}\label{line:fuzzy-matching}
    \State $spans.insert(s)$
\EndFor
\EndIf
\State \Return{$spans$}
\EndFunction
\State
\Function{Preprocess}{$str$: string}
    \State $str \gets Lowercase(str)$
    \State $str \gets RemovePunctuation(str)$
    \State \Return $str$
\EndFunction
\end{algorithmic}
\caption{\label{alg:kelm-matching}
KELM-Wikidata matching algorithm that finds spans in KELM sentences corresponding to Wikidata KG entities. $a \subset b$ means that $a$ is a substring of $b$. $*$ represents any string.
}
\end{algorithm}

We don't try to match \textit{relation} part of triples, because it could be represented in many different forms. For example, the triple (\texttt{Pulp Fiction}, \texttt{cast member}, \texttt{John Travolta}) could be represented as "\texttt{John Travolta was an actor in Pulp Fiction}", "\texttt{John Travolta starred in Pulp Fiction}", "\texttt{John Travolta played Vincent Vega in Pulp Fiction}", etc., and there is no way to robustly align a relation to all possible surface forms.

\section{Dataset}

Wikidata \cite{wikidata} was released under the Creative Commons CC0 License. KELM \cite{agarwal-etal-2021-knowledge} was released under the Creative Commons CC BY-SA 2.0 License. NaturalQuestions \cite{kwiatkowski-etal-2019-natural} and WikiHop \citep{welbl-etal-2018-constructing} were released under Creative Commons CC BY-SA 3.0 License. MetaQA \citep{zhang2017variational} was released under Creative Commons CC BY-ND 3.0 License. C4 \cite{t5} and TriviaQA \cite{joshi-etal-2017-triviaqa} were released under Apache-2.0 License. WikiMovies \cite{miller-etal-2016-key} was released under MIT License. FreebaseQA \citep{jiang-etal-2019-freebaseqa}\footnote{\url{https://github.com/kelvin-jiang/FreebaseQA}} was released without a license.

\label{sec:dataset-sizes}
\begin{table}[!h]
\centering
\begin{adjustbox}{center}
{\footnotesize
\begin{tabular}{lccc}
\hline
                 & \textbf{train} & \textbf{dev} & \textbf{test} \\
\hline
FreebaseQA       & $20,358$          & $3,994$         & $3,996$ \\
WikiHop          & $39,364$          & $4,374$         & $5,129$ \\
TQA              & $78,785$          & $8,837$        & $11,313$ \\
TQA-matched      & $20,948$          & $2,289$         & $3,064$ \\
NQ               & $79,168$          & $8,757$        & $3,610$  \\
NQ-matched       & $10,487$          & $1,160$         & $462$   \\
MetaQA-1hop      & $96,106$          & $9,992$         & $9,947$  \\
MetaQA-2hop      & $118,980$         & $14,872$        & $14,872$ \\
MetaQA-3hop      & $114,196$         & $14,274$        & $14,274$ \\
\hline
\end{tabular}}
\end{adjustbox}
\caption{\label{table:dataset-sizes}
Dataset sizes for the closed-book QA tasks. TQA and NQ stands for TriviaQA and NaturalQuestions, respectively. *-matched are the selected dataset with the Wikidata KG answerable questions, and the KG alignment details can be found in \Cref{sec:wikidata-answerable-qa}.}
\end{table}

\end{document}